\documentclass[conference]{IEEEtran}
\IEEEoverridecommandlockouts
\usepackage{cite}
\usepackage{amsmath,amssymb,amsfonts}
\usepackage{graphicx}
\usepackage{subcaption}
\usepackage{textcomp}
\usepackage{xcolor}
\usepackage{fdsymbol}
\usepackage{bbding}
\usepackage{url}
\usepackage{multirow}
\usepackage[normalem]{ulem}
\usepackage{algorithm}
\usepackage{algorithmic}
\usepackage{amsmath}
\usepackage{booktabs}
\usepackage{eso-pic} 

\def\BibTeX{{\rm B\kern-.05em{\sc i\kern-.025em b}\kern-.08em
    T\kern-.1667em\lower.7ex\hbox{E}\kern-.125em}}

\usepackage[square,numbers,sort]{natbib}
\begin{document}
\AddToShipoutPictureBG*{%
  \AtPageUpperLeft{%
    \setlength\unitlength{1in}%
    \hspace*{\dimexpr0.5\paperwidth\relax}
    \makebox(0,-0.75)[c]{\textit{OCEANS 2025 - Great Lakes}}
}}

\title{Robust Underwater Localization of Buoyancy Driven $\mu$Floats using Acoustic Time-of-Flight Measurements

\thanks{Work funded by the National Science Foundation Physical Oceanography Award \#2319495}
}

\author{\IEEEauthorblockN{
Murad Mehrab Abrar}
\IEEEauthorblockA{\textit{Applied Physics Laboratory} \\
\textit{University of Washington}\\
Seattle, WA, USA \\
\url{mabrar@uw.edu}}
\and
\IEEEauthorblockN{Trevor W. Harrison}
\IEEEauthorblockA{\textit{Applied Physics Laboratory} \\
\textit{University of Washington}\\
Seattle, WA, USA \\
\url{twharr@uw.edu}}
}

\maketitle

\begin{abstract}

Accurate underwater localization remains a challenge for inexpensive autonomous platforms that require high-frequency position updates. In this paper, we present a robust, low-cost localization pipeline for buoyancy-driven $\mu$Floats operating in coastal waters. We build upon previous work by introducing a bidirectional acoustic Time-of-Flight (ToF) localization framework, which incorporates both float-to-buoy and buoy-to-float transmissions, thereby increasing the number of usable measurements. The method integrates nonlinear trilateration with a filtering of computed position estimates based on geometric cost and Cramér-Rao Lower Bounds (CRLB). This approach removes outliers caused by multipath effects and other acoustic errors from the ToF estimation and improves localization robustness without relying on heavy smoothing. We validate the framework in two field deployments in Puget Sound, Washington, USA. The localization pipeline achieves median positioning errors below 4 m relative to GPS positions. The filtering technique shows a reduction in mean error from 139.29 m to 12.07 m, and improved alignment of trajectories with GPS paths. Additionally, we demonstrate a Time-Difference-of-Arrival (TDoA) localization for unrecovered floats that were transmitting during the experiment. Range-based acoustic localization techniques are widely used and generally agnostic to hardware---this work aims to maximize their utility by improving positioning frequency and robustness through careful algorithmic design. 

\end{abstract}


\section{Introduction}

Oceanographic floats play a vital role providing climate-relevant ocean data. As a category, floats are simple autonomous robots that actuate vertically in the water column by controlling their buoyancy, but otherwise drift passively with the surrounding flow, following a near-Lagrangian trajectory. Deployment of a fleet (or ``swarm") of floats that drift with ocean currents has proven successful in observing circulation and climate variability in the global ocean with the Argo program \cite{jayne2017argo}. Argo-style floats collect high-quality, long-term measurements of temperature, salinity, and pressure in the upper 2,000 meters of the ocean \cite{roemmich2009argo}. 

When considering the in situ sensing needs and modeling of the coastal ocean \cite{arkema2015embedding, wilkin2017advancing, liu2015introduction}, deploying swarms of floats is a conceptually attractive approach as well. However, Argo-style floats are designed for global-scale monitoring in the open ocean, and consequently ill-suited to coastal or nearshore environments \cite{siiria2019applying}. The highly variable hydrodynamics, shallow water, and strong density gradients present in coastal waters necessitate high sampling rates, fast and tight buoyancy control,  and frequent location updates to sample effectively in this environment. 

To overcome the challenges of coastal settings, the $\mu$Float (pronounced microfloat) offers a promising alternative. The $\mu$Float is a compact and cost-effective variant of traditional floats that is designed specifically for distributed sensing and sampling in shallow coastal waters \cite{Harrison2023benchmark}. At present, the system is suitable for short-duration missions (typically less than a day) to capture rapidly evolving phenomena in dynamic coastal environments. 

Frequent and accurate tracking of the float trajectory while underwater provides direct insight on water circulation and evolution. Since GPS is unavailable below the surface, underwater tracking is typically achieved via acoustic methods. Coastal environments are particularly challenging for acoustics, as multipath propagation errors (acoustic reflections off the seafloor or sea surface), noise (vessel or environmental), and strong sound speed gradients may all act to reduce the robustness of acoustic communications and accuracy of derived localization measurements.

Motivated by these challenges, this paper presents a robust and comprehensive pipeline to localize inexpensive robots like $\mu$Floats--- \textit{which require high-frequency and robust localization but are equipped with relatively simple acoustic modems}
--- using the acoustic Time-of-Flight (ToF) measurements. Our key contributions include:

\begin{itemize}
    \item Development of a bidirectional ToF-based localization algorithm to maximize measurement density and improve localization robustness.


    \item Integration of a filtering approach that checks both cost function value (sum of squared residuals) and two-dimensional (2D) Cramér-Rao Lower Bounds (CRLB) on position estimates to remove outliers.

    \item Presentation and analysis of two field deployments conducted in Puget Sound, Washington, USA, to validate the performance of the localization pipeline in real-world coastal environments.
\end{itemize}

\section{Background and Related Work}

\subsection{Overview of the $\mu$Float System}

The $\mu$Float system consists of a swarm of (i) $\mu$floats, (ii) Surface Localization Buoys (SLBs), and (iii) moored Underwater Localization Buoys (ULBs). The $\mu$Floats and SLBs are equipped with GPS, and all three devices are capable of exchanging acoustic messages with each other via a low-cost ($\$$600) Succorfish Delphis v3 acoustic nanomodem. The modem transmits messages on a carrier frequency band of 24-32 kHz. They support a maximum data rate of 463 bit/s, a source level of 168 dB referenced to 1 $\mu$Pa at 1 m, and a nominal communication range of approximately 2 km \cite{fenucci2018development}. The nanomodems meet the rapid communication requirements at significantly lower cost than conventional acoustic modems. It provides full duplex (both transmission and reception) capabilities, but cannot send and receive pings simultaneously and parse overlapping messages; instead, it alternates between transmit and receive modes. Therefore, a Time Division Multiple Access (TDMA) protocol is used to schedule and coordinate localization pings across the array, with a common messaging schedule provided to all floats and SLBs prior to deployment. Our implementation utilized one-way broadcast ``pings" that are encoded with the device source address (the ``Frame Beacon" command type in Delphis Firmware v1.6.0). All modems within the broadcast range record and timestamp the received messages. Floats are pre-programmed with a schedule of dive depths and durations. The float extends a solid piston in and out of its housing to adjust its density, thus controlling depth while drifting with local water currents. 
For more details on float hardware-software architecture and depth control techniques, see \cite{Harrison2023benchmark}.

\subsection{Related Work in Underwater Localization}

Accurate underwater localization techniques include acoustic trilateration/ multilateration using Long Baseline (LBL) and Short Baseline (SBL) systems, as well as range and bearing-based Ultra-Short Baseline (USBL) methods \cite{stutters2008navigation, kinsey2006survey}. While these systems can achieve sub-meter accuracy, they are time-intensive to set up and rely on expensive infrastructure, making them impractical for low-cost deployments.

Range-based methods such as Time-of-Arrival (ToA), Time-Difference-of-Arrival (TDoA), and Angle-of-Arrival (AoA) have also been widely used \cite{foy2007position}. 
Among these, AoA requires directional antennas or hydrophone arrays to estimate angles, which increases system complexity and cost. In general, range-based methods are prone to error due to (1) the assumption of direct line-of-sight (LOS) between transmitter and receiver, and (2) significant multipath effects in underwater environments \cite{li2016contributed}. That said, for small and inexpensive platforms like $\mu$Floats, TOA and TDOA are preferable techniques due to their compatibility with inexpensive acoustic hardware.

\subsection{Prior work on $\mu$Float Localization}
Prior work has demonstrated successful $\mu$Float localization using acoustic nanomodems and unidirectional ToF measurements from ToA, where the GPS-tracked SLBs transmitted pings and nearby floats received during each round-robin cycle \cite{Harrison2023benchmark}.
To estimate 2D float positions, ToF measurements from at least three SLBs within a cycle were trilaterated using a least-squares fit \cite{norrdine2012algebraic}. The localization method was previously chosen as the position update rate scaled inversely with the number of SLBs (5) rather than the number of floats (up to 20), but suffered from error resulting from SLB movement during the round-robin cycle and unfavorable acoustic gradients \cite{Harrison2023benchmark}. The resulting localizations were noisy and intermittent, requiring smoothing (e.g., MATLAB's \texttt{rloess}) over 60-240~s windows to yield physically realistic float trajectories \cite{Harrison2023benchmark}.

This paper builds upon the existing framework by introducing a bidirectional ToF localization algorithm that utilizes both SLB-to-float and float-to-SLB transmissions. This increases flexibility of messaging and robustness to the acoustic environment, improving overall system performance while maintaining sample frequency. In contrast to previous reliance on smoothing, we introduce an outlier-rejection-based approach that checks both optimization cost and the CRLBs to remove only erroneous position estimates without suppressing high-frequency motion in high-quality estimates. 


\section{Localization Framework}

The localization pipeline flows through the following steps: (A) ping matching and range calculation, (C) depth compensation, (D) initial outlier removal, (E) ping grouping, (F) group-based nonlinear trilateration, (G) uncertainty estimation, and (H) final position filtering. To clarify, this pipeline is intended for post-processing once all devices have been recovered. In the case of unrecovered floats, the pipeline includes a fallback TDoA-based algorithm, as it does not rely on knowledge of the ping transmit time from the float.

\subsection{Bidirectional Ping Match and Time-of-Flight Computation}

Each acoustic ping contains only the sender's device ID and is transmitted according to a predefined TDMA schedule. All transmissions and receptions are timestamped with absolute time (UTC) on the local device. The localization process considers two directions of acoustic exchanges:

\begin{itemize}
    \item \textbf{Float-to-SLB (Uplink):} 
    The float transmits and nearby SLBs receive the ping. Since a ping sent by a float can be received by several SLBs nearly simultaneously if they are within acoustic range, uplink generally yields more reliable localization with lower uncertainty. These messages also permit TDoA-based localizations. 

    \item \textbf{SLB-to-Float (Downlink):} SLBs transmit and float receives. Here, multiple SLBs must transmit pings within a tightly synchronized time window so that the relative drift of the SLBs and float are small, as this adds uncertainty to localization, especially when drifting in high speed currents.

\end{itemize}

We use both uplink and downlink directions for float localization. Ping pairs are found by matching sent ID with receive ID across platforms where the difference between send and receive time is less than 1.5 seconds (effective range under 2 km, the expected limit of the modem). Each pair is 
converted into a ToF measurement, accounting for software processing delays. The resulting ToF values are then translated into acoustic distances (or ranges) using a measured sound speed, $c$, as ${d_{\text{acoustic}} = c \cdot ToF}$. While there are also moored ULBs in the system, inclusion of their data remains future work.


\subsection{Depth Compensation and Horizontal Distance Calculation}

Floats depth is already know from the pressure sensor, so the localization algorithm must only estimate the 2D underwater position of a target float. As such, acoustic distance must be converted to horizontal distances between the float and each SLB. To compute the horizontal distances ($d_\text{horizontal}$), each ToF-based acoustic distance ($d_\text{acoustic}$) is adjusted for the relative depth offset between the float and the SLBs according to:
\noindent
\begin{equation}
    d_{\text{horizontal}} = \sqrt{d_{\text{acoustic}}^2 - (z_{\text{Float}} - z_{\text{SLB}})^2}
    \label{eq1}
\end{equation}

where $z_{\text{Float}}$ is the depth of the target float as measured by its pressure sensor and $z_{\text{SLB}}$ is a constant value of 3 m, based on the SLB hardware. We note that this method assumes a geometric ray path, ignoring acoustic refraction.


\subsection{Outlier Removal from Range Measurements}
To ensure realistic results, the pipeline filters out physically impossible range measurements and checks the horizontal distances among the devices. Range measurements are filtered in two stages:

\begin{itemize}
    \item \textbf{Invalid Acoustic Distances:} Acoustic distances less than the depths of the float are identified and discarded.
    
    \item \textbf{Inconsistency in Horizontal Distances:} Non-physical changes in horizontal distances with time are identified by a velocity check (within 0.8 m/s) and discarded.
\end{itemize}

\subsection{Grouping Pings for Joint Localization}

Using the filtered horizontal distances, we build groups of pings sent within a fixed sliding time window (5 seconds) and assign a group ID. Only groups with three or more unique SLBs are used for trilateration, though SLB can be either transmitter or receiver. That is, groups can contain both uplink and downlink pings as long as send times occur within the time window. This bidirectional strategy increases the number of usable measurements and improves localization robustness, especially in sparse data conditions \cite{jiang2007asymmetric}.

\subsection{Nonlinear Least Squares Trilateration}

We then apply a nonlinear least-squares trilateration model to each ping group to estimate an absolute position of the target float. The steps are as follows:


\subsubsection{Initial Guess Interpolation between Dive and Rise}

For each group, an initial guess of the horizontal position is required for the trilateration. This initial guess is calculated by a time-based linear interpolation between the dive and rise GPS locations of the float. All positions (SLB GPS and initial guess of the float position) are converted to a local North-East-Down (NED) frame using a nearby fixed geo-reference point (MATLAB's \texttt{geodetic2ned}). This converts from latitude and longitude into an equal-length reference frame. 

\subsubsection{Nonlinear Optimization Formulation}

Given $N$ horizontal acoustic distances $\{d_{horizontal}\}$ and the corresponding SLB positions $\{\mathbf{p}_i = (x_i, y_i)\}$, we estimate the float position $\mathbf{p} = (x, y)$ by solving the following minimization:

\begin{equation}
    \min_{\mathbf{p}} \sum_{i=1}^{N} \left( \| \mathbf{p} - \mathbf{p}_i \| - d_{horizontal} \right)^2
    \label{eq:optimization}
\end{equation}

This is solved using MATLAB's \texttt{fminsearch} that yields an estimated float position $\mathbf{p}^* = (x^*, y^*)$ in the NED frame.

\subsection{CRLB-based Uncertainty Estimation}

To quantify the theoretical uncertainty of each estimated float position, we compute the Cramér--Rao Lower Bounds using the Fisher Information Matrix (FIM). The CRLB establishes a theoretical lower bound on the variance of any unbiased estimator, meaning it represents the theoretical best-case precision achievable for a given set of measurements and a specific parametric model. In the context of underwater acoustic positioning, the CRLB serves as a theoretical benchmark to evaluate localization performance where ground truth is not available \cite{zhang2020efficient, li2016contributed, van1968detection, kay1993fundamentals}. 
 
The FIM quantifies the amount of information that a set of observable random variables (e.g., the acoustic range measurements) carries about an unknown parameter or set of parameters (e.g., the float's position) \cite{kay1993fundamentals}. A larger FIM value indicates that the measurements are highly informative, leading to a smaller CRLB and a potentially more precise position estimate. For each trilateration group with $N$ valid pings, the FIM, $\mathbf{F}$, is computed as:

\begin{equation*}
    \mathbf{F} = \frac{1}{\sigma^2} \sum_{i=1}^{N} \frac{1}{r_i^2}
\begin{bmatrix}
(x^* - x_i)^2 & (x^* - x_i)(y^* - y_i) \\
(x^* - x_i)(y^* - y_i) & (y^* - y_i)^2
\end{bmatrix}
\end{equation*}

where $(x_i, y_i)$ are the known SLB positions, $(x^*, y^*)$ is the estimated float position, $r_i = \| \mathbf{p}^* - \mathbf{p}_i \|$ is the distance between the estimated float position and the $i\textsuperscript{$th$}$ SLB, and $\sigma^2$ is the variance of the acoustic measurement noise. We assume a fixed noise variance $\sigma^2 = 9 \, \text{m}^2$ based on empirical calibration in Puget Sound. If the FIM is invertible, the covariance matrix of the estimator is given by:

\begin{equation*}
    \text{Cov}(\hat{\mathbf{p}}) = \mathbf{F}^{-1}
\end{equation*}

The diagonal entries of this matrix provide the lower bounds on the variances of $x$ and $y$ estimates, from which the standard deviations $\sigma_x$ and $\sigma_y$ are extracted. These values are used to draw uncertainty ellipses centered at the estimated float positions. The size and orientation of each ellipse reflect the confidence of the estimate and the geometry of the surrounding SLBs. Groups with ill-conditioned FIMs 
are not considered for position estimates.

\subsection{Filtering and Final Validation}\label{section: filter}

As a final step, we remove outlier position estimates by applying two criteria:

\begin{itemize}
    \item \textbf{Trilateration Residual Cost Check:} The nonlinear least squares optimization outputs the residual cost after the minimization process has completed. This residual reflects how well the estimated float position explains the observed range measurements. Only position estimates where this cost is below a predefined threshold $\tau_{\text{cost}}$ are considered acceptable. A high cost indicates inconsistency among the range observations or a poor geometric configuration of SLBs.
    
    \item \textbf{CRLB Uncertainity Check:} The CRLB-derived standard deviations $\sigma_x$ and $\sigma_y$ quantify the theoretical positional uncertainty in the $x$ and $y$ directions, respectively. These values reflect both the measurement noise and the spatial distribution of SLBs. Position estimates are retained only if both uncertainties are below a predefined threshold $\tau_{\text{crlb}}$.
  
\end{itemize}

Formally, the filtering condition is:
\noindent
\begin{equation*}
    \text{Cost} \leq \tau_{\text{cost}} \quad \text{and} \quad \text{Positional Uncertainty,} \hspace{2pt} (\text{$\sigma$}_x, \text{$\sigma$}_y) \leq \tau_{\text{crlb}}
    \label{eq:filter_constraints}
\end{equation*}

Position estimates that pass both filters are converted back to geodetic coordinates using the inverse transformation (MATLAB's \texttt{ned2geodetic}).

\subsection{Time Difference of Arrival (TDoA) Localization}

TDoA works as a fallback method in the localization pipeline to estimate the position of actively transmitting floats whose data are unavailable (e.g., lost floats). Unlike ToF-based methods, TDoA relies solely on the relative arrival times of pings recorded by multiple SLBs. Pings received within a predefined time window are grouped. The SLB with the earliest arrival is treated as the reference. The arrival time differences between each SLB and the reference are multiplied by the speed of sound to obtain range differences, which form the basis for localization. A 2D cost function is constructed to minimize residuals between the observed and modeled range differences. The float position $\mathbf{p} = (x, y)$ is estimated by solving a nonlinear optimization problem, with the initial guess set to the centroid of the SLB positions. Given $N$ SLBs with known positions $\{\mathbf{p}_i = (x_i, y_i)\}$, and assuming $\mathbf{p}_1$ is the reference SLB, the range differences $\{\Delta d_i\}$ for $i = 2, \dots, N$ are defined as:

\begin{equation*}
    \Delta d_i = c \cdot (t_i - t_1)
\end{equation*}

where $t_i$ is the signal arrival time at SLB $i$, and $c$ is the speed of sound underwater. The position is then estimated by minimizing the following equation:

\begin{equation*}
    \min_{\mathbf{p}} \sum_{i=2}^{N} \left( \| \mathbf{p} - \mathbf{p}_i \| - \| \mathbf{p} - \mathbf{p}_1 \| - \Delta d_i \right)^2
\end{equation*}

The resulting position estimates $\mathbf{p}^* = (x^*, y^*)$ are filtered using the same outlier rejection scheme described in G. 

The localization pipeline is summarized in Algorithm \ref{alg:tof_localization}.

\begin{algorithm}[H]
\caption{Time of Flight-based Localization of $\mu$Floats}\label{alg:tof_localization}
\begin{algorithmic}[1]
\REQUIRE Table of all acoustic pings, underwater sound speed $c$, software processing delay $\delta$, transmit-receive time window $t_w$, grouping time window $t_g$

\STATE Extract exchanged pings between floats and SLBs

\hspace{-20pt} \textbf{Match sent $\rightarrow$ received pings:}

    \FORALL{sent pings}
        \STATE Find the ping recorded by receivers within $t_w$
        \STATE Compute Time-of-Flights, $ToF = t_{recv} - t_{send} - \delta$        \STATE Compute acoustic distances, ${d_{\text{acoustic}} = c \cdot ToF}$
        \STATE Compute horizontal distances, $d_{\text{horizontal}}$ \quad [Eq.~\ref{eq1}]
        \STATE Record entry
    \ENDFOR

\STATE Filter entries with invalid distances, depth, and velocites.

\STATE Group pings based on sent time within $t_g$

\hspace{-20pt} \textbf{Perform Trilateration:}

\FORALL{Group with $\geq 3$ SLBs}
    \STATE Interpolate geodetic initial guess from float dive to rise straight line
    \STATE Convert geodetic initial guess to NED initial guess $(x_0, y_0)$
    \STATE Solve nonlinear least squares trilateration \quad [Eq.~\ref{eq:optimization}]
    \STATE Get estimated NED position $(x^*, y^*)$
    \STATE Compute CRLB from Fisher Information Matrix
    \STATE Apply cost and CRLB filter on $(x^*, y^*)$ $\rightarrow$ $(x, y)$
    \STATE Convert filtered $(x, y)$ to [longitude, latitude]
\ENDFOR


\RETURN Estimated positions in [longitude, latitude] with residual cost from optimization and CRLB bounds.
\end{algorithmic}
\end{algorithm}

\begin{figure*}[t]
    \centering
    \includegraphics[width=0.85\linewidth]{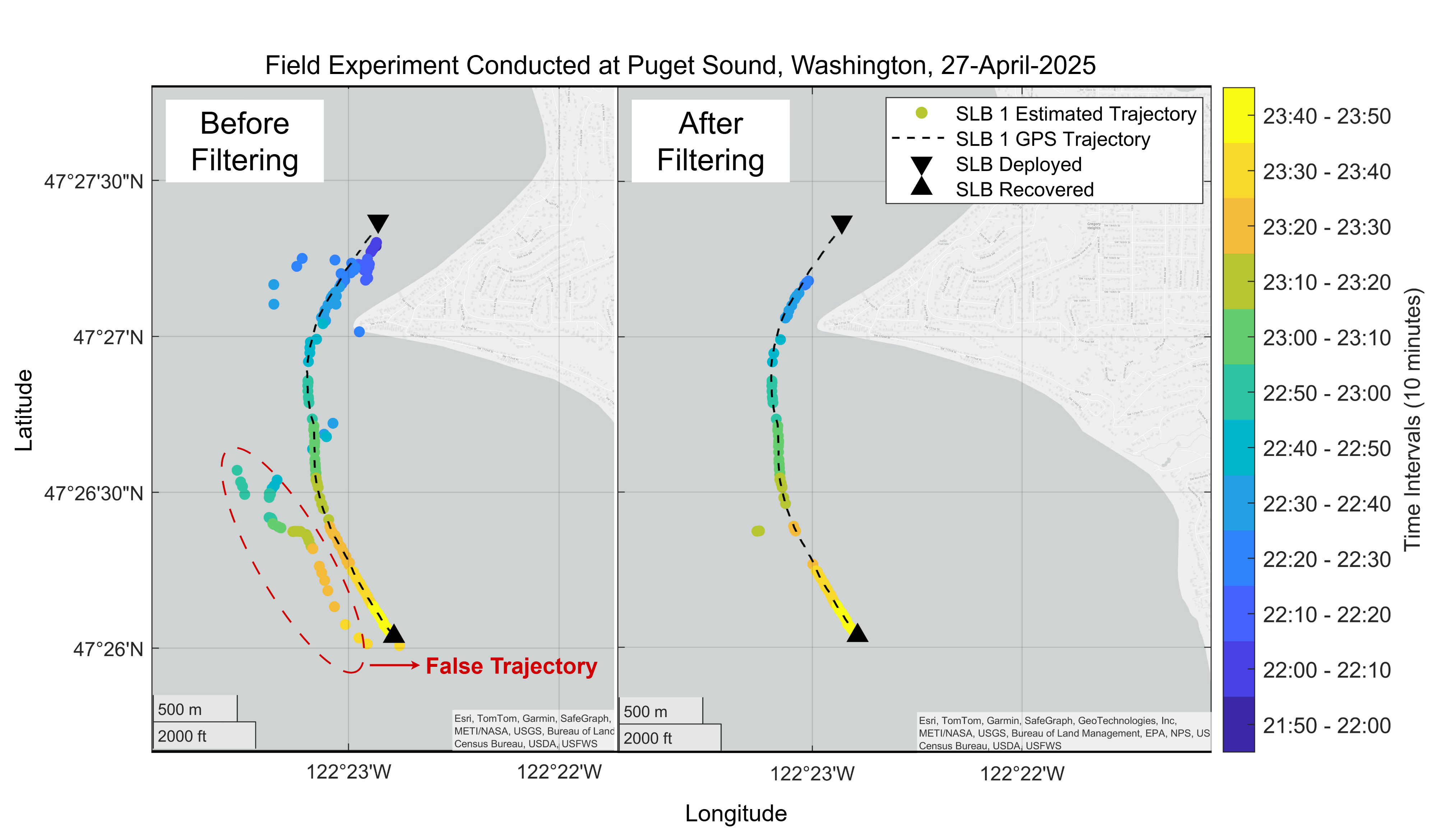}
    \caption{SLB 1 localization before (left) and after (right) applying the CRLB and cost filter. The unfiltered trajectory contains some outliers and a false path. After filtering, these outliers are removed and the estimated trajectory closely aligns with the true GPS path.}
    \label{fig:slb1_filter_comparison}
\end{figure*}

\begin{figure*}[t]
    \centering
    \begin{subfigure}[t]{0.81\columnwidth}
        \centering  \includegraphics[width=\linewidth]{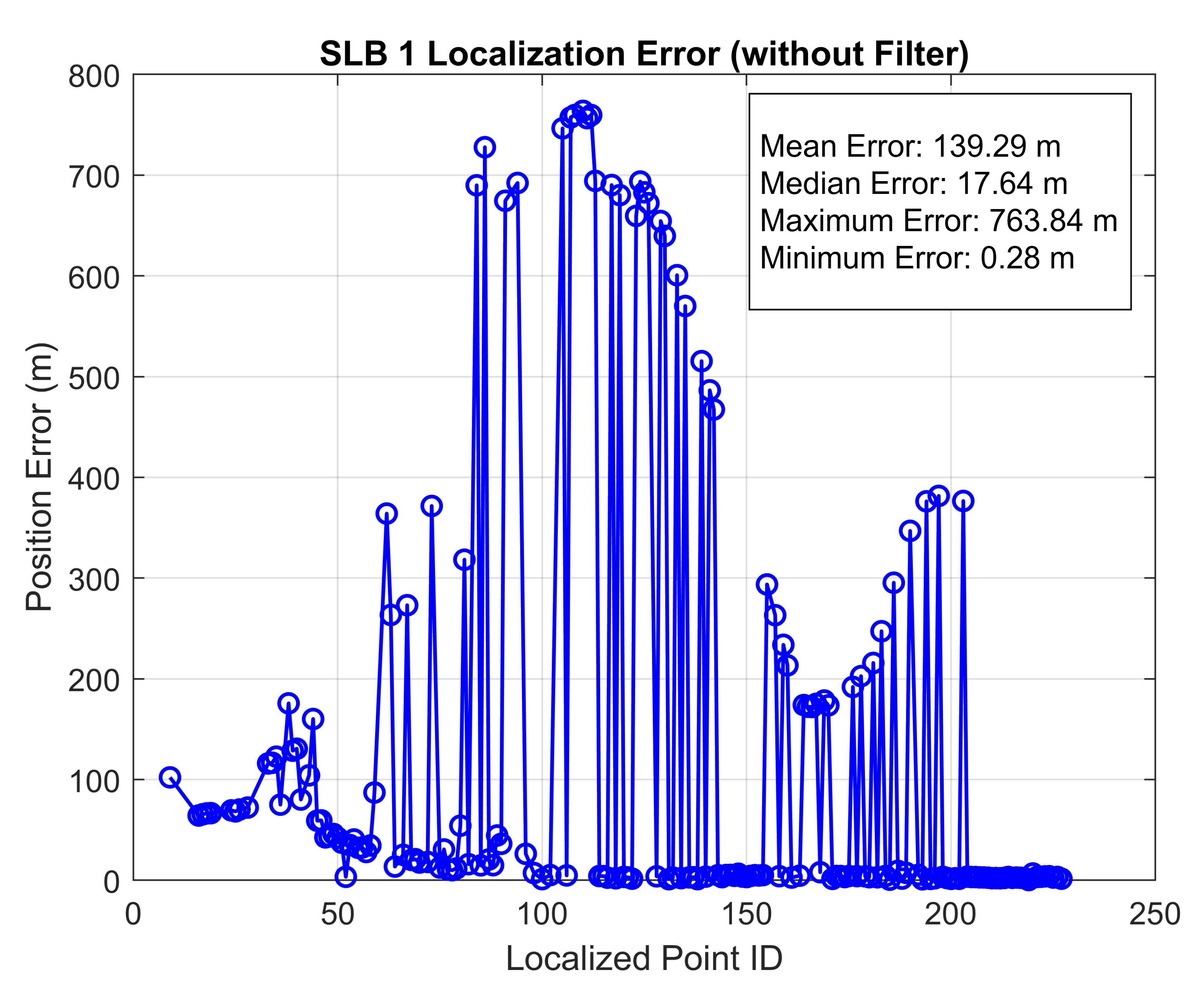}
        \caption{SLB 1 localization error before filtering}
        \label{fig:slb1_before_filter}
    \end{subfigure}
    \hspace{1.6cm}
    \begin{subfigure}[t]{0.81\columnwidth}
        \centering  \includegraphics[width=\linewidth]{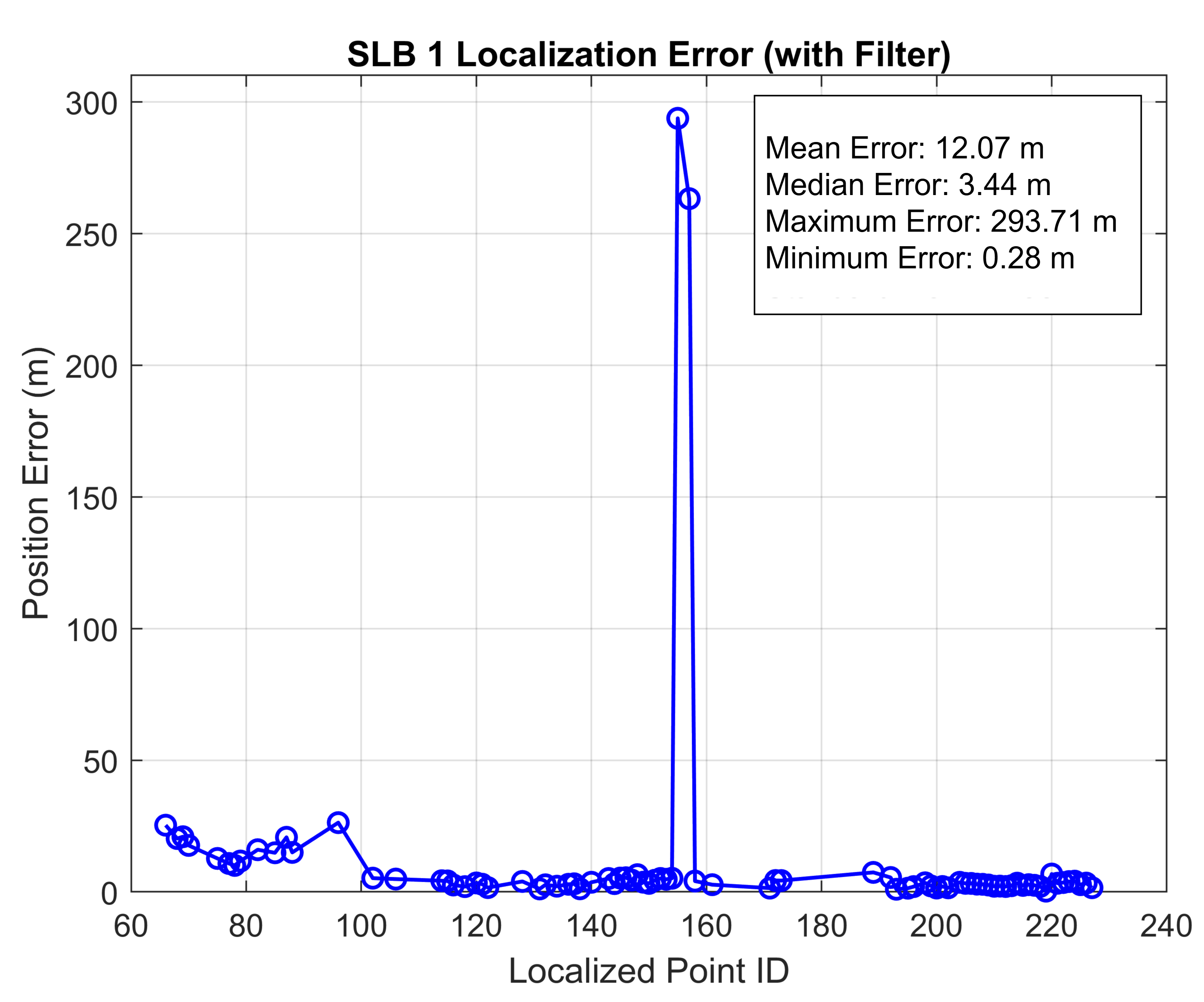}
        \caption{SLB 1 localization error after filtering}
        \label{fig:slb1_after_filter}
    \end{subfigure}
    \caption{SLB 1 localization error before (a) and after (b) applying filter. Without filtering, the error distribution shows multiple extreme outliers. After filtering, the majority of outliers are removed.}
    \label{fig: localization error comparison}
\end{figure*}

\begin{figure*}[t!]
    \centering       \includegraphics[width=0.74\linewidth]{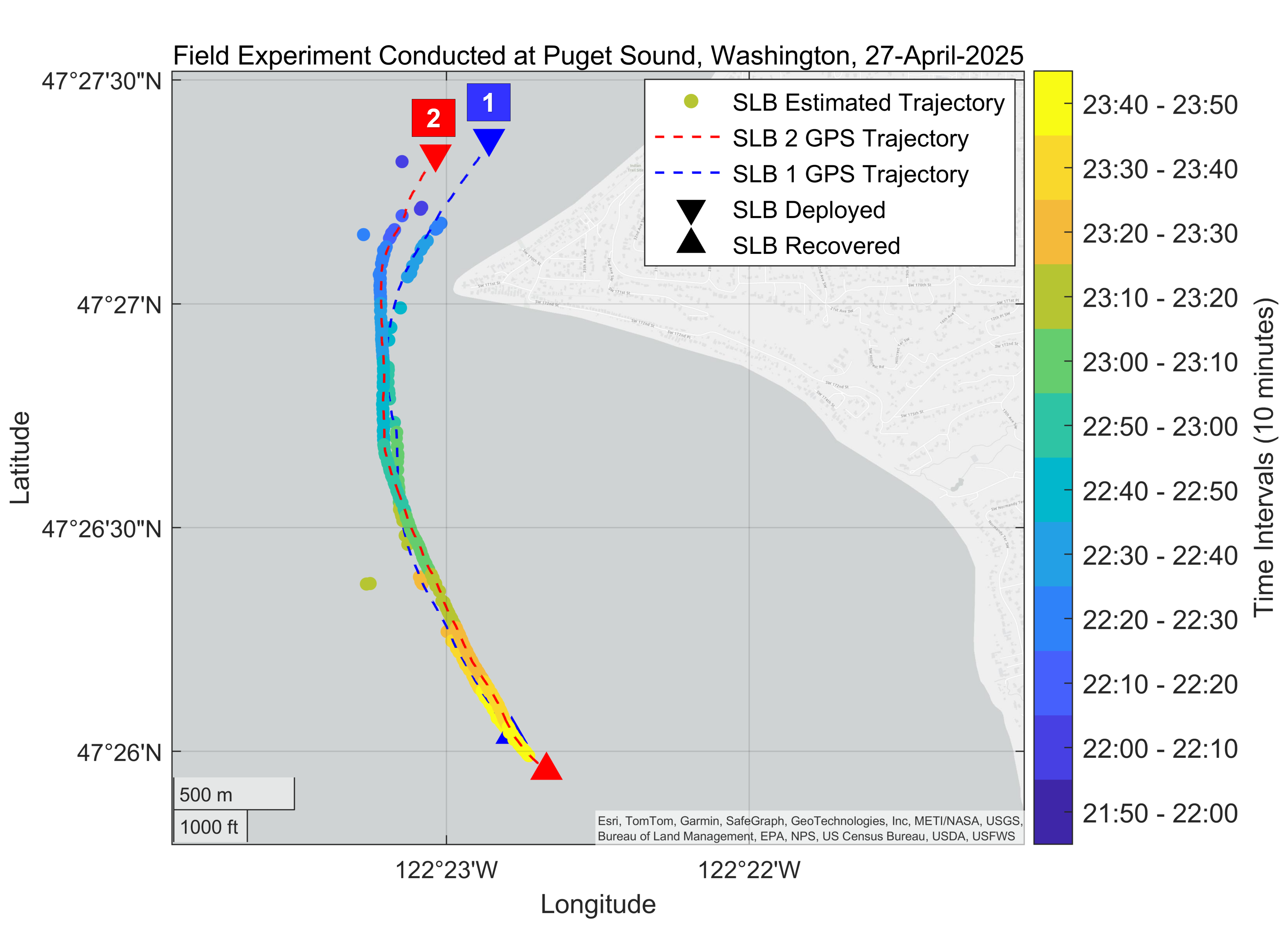}
        \caption{Filtered trajectories of SLB 1 and SLB 2 overlaid on GPS path.}\label{slb_localization}
\end{figure*}

\section{Implementation and Results}

\subsection{Field Deployments}

Two field experiments were conducted in Puget Sound, Washington, USA, from April 26--27, 2025. Floats and SLBs were deployed on the upstream side of a headland (Three Tree Point) during the flood tide, with currents flowing south around the point at 0.75 to 1.5 m/s. The bathymetry of the site is highly variable, ranging from 30 to 180 meters within the sampled region \cite{pawlak2003observations}. Each experiment deployed an array of SLBs and $\mu$Floats, all equipped with acoustic modems operating under a synchronized TDMA schedule.

\begin{itemize}
    \item \textbf{Field Deployment I (April 26, 2025):} One $\mu$Float (ID 38) and five SLBs (IDs 1--5) were deployed for approximately 2 hours.
    \item \textbf{Field Deployment II (April 27, 2025):} Thirteen $\mu$Floats (IDs: 7, 9, 22, 23, 24, 32, 33, 35, 36, 37, 39, 40, and 42) and five SLBs were deployed for approximately 1 hour.
\end{itemize}

\subsection{Localization Performance Evaluation}
To evaluate the performance of the localization algorithm, we treated SLBs as target floats and compared ToF-estimated SLB positions against their known GPS positions. While some depth-dependent performance effects are omitted by this assessment, SLB to SLB localizations serve as a convenient proxy for evaluation, as the acoustic hardware is identical between SLBs and $\mu$Floats and the SLB's GPS data provides an effective reference. Due to a better SLB geometry on April 27, 2025, data from Field Deployment II was used to evaluate SLB-SLB localization accuracy using ToF. SLB 1 and SLB 2 had sufficient pings communicated with other SLBs, therefore, the localization algorithm was evaluated on these two SLBs.

\subsubsection{Filtering}

To remove the outliers, a filter was used that restricts the cost function value and the CRLB-defined position uncertainty, as detailed in Section~\ref{section: filter}. Based on manual tuning, the following filter delivered reliable results for SLB-SLB localization:
\noindent
\begin{equation*}
    \text{Cost} \leq 50 \quad \text{and} \quad \text{Positional Uncertainty,} \hspace{2pt} (\text{$\sigma$}_x, \text{$\sigma$}_y) \leq 10 \, \text{m}
    \label{eq:filter_constraints}
\end{equation*}

As shown in Fig.~\ref{fig:slb1_filter_comparison}, the filter removes nearly all outliers, including the false trajectory observed in the lower left of the unfiltered trajectory that was likely the result of multipath effects on a particular SLB-to-SLB pair. 

Without filtering, the position error (computed as the distance between estimated position and GPS position) has numerous and significant outliers resulting in a high mean error of 139.29 m and median error of 17.64 m (Fig.~\ref{fig: localization error comparison}a). After filtering, these outliers are largely removed, and the error distribution becomes more compact, with a reduced mean error of 12.07 m  (Fig.~\ref{fig:slb1_after_filter}). 
Although the filtering process does remove some good estimates (position error less than 4 m), its overall impact improves localization agreement with the GPS ground truth. Therefore, this filter is used in the subsequent localization processing.

\begin{table}[!h]
    \centering
    \caption{Statistical Summary of Localization Errors for SLB 1 and SLB 2}
    \label{tab:localization_error}
    \begin{tabular}{lcc}
        \toprule
        \textbf{Metric} & \textbf{SLB 1} & \textbf{SLB 2} \\
        \midrule
        Mean Error (m)     & 12.07  & 6.71   \\
        Median Error (m)   & 3.44   & 3.67   \\
        Maximum Error (m)  & 293.71 & 165.79 \\
        Minimum Error (m)  & 0.28   & 0.41   \\
        \bottomrule
    \end{tabular}
\end{table}


\subsubsection{SLB Trajectory and Error Visualization}

Fig.~\ref{slb_localization} illustrates the final filtered trajectories of SLB 1 and SLB 2 overlaid on their true GPS paths. We can see that the localized tracks follow the GPS trajectories in both SLBs, with most outliers removed. This visual agreement is confirmed by position error metrics summarized in Table~\ref{tab:localization_error}. Notably, both SLBs achieve median localization errors under 4 m (3.44 m and 3.67 m, respectively), indicating that at least half of the localization estimates are within 4 m of the GPS position. 

Although the mean error for SLB 1 is higher due to two large deviations, both exceeding 250 m (visible in Fig.~\ref{fig:slb1_after_filter}), the low median value suggests that these are outliers rather than representative errors. These outliers were not removed by the applied filtering technique. In contrast, SLB 2 shows lower mean and median values, which reflect more consistent localization performance overall.

\begin{figure*}[t!]
    \centering
    \begin{subfigure}[t]{0.45\textwidth}
        \centering
        \includegraphics[width=\linewidth]{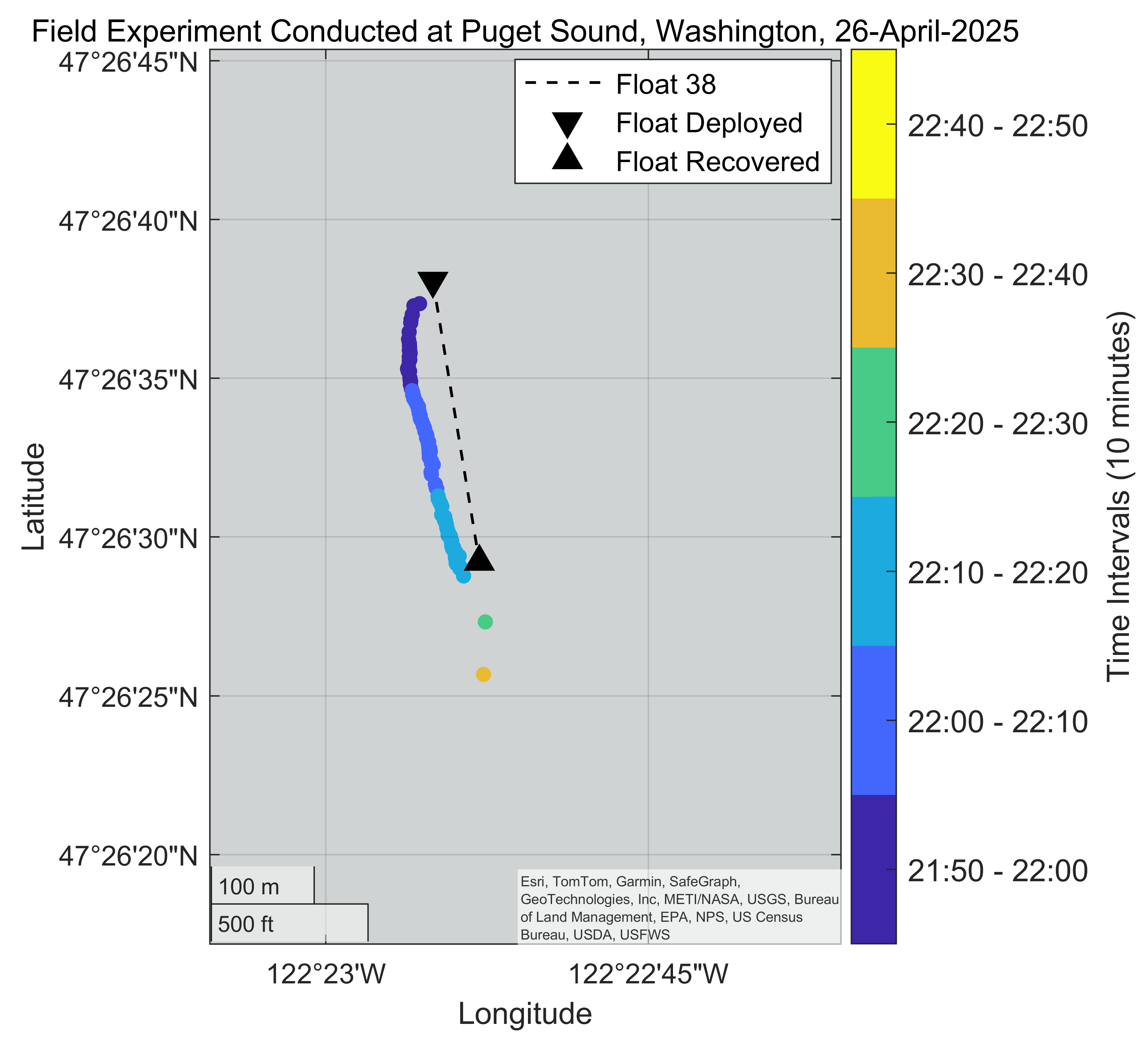}
        \caption{Localization of Float 38 from Field Deployment I}
        \label{fig:float38}
    \end{subfigure}
    \hfill
    \begin{subfigure}[t]{0.525\textwidth}
        \centering
        \includegraphics[width=\linewidth]{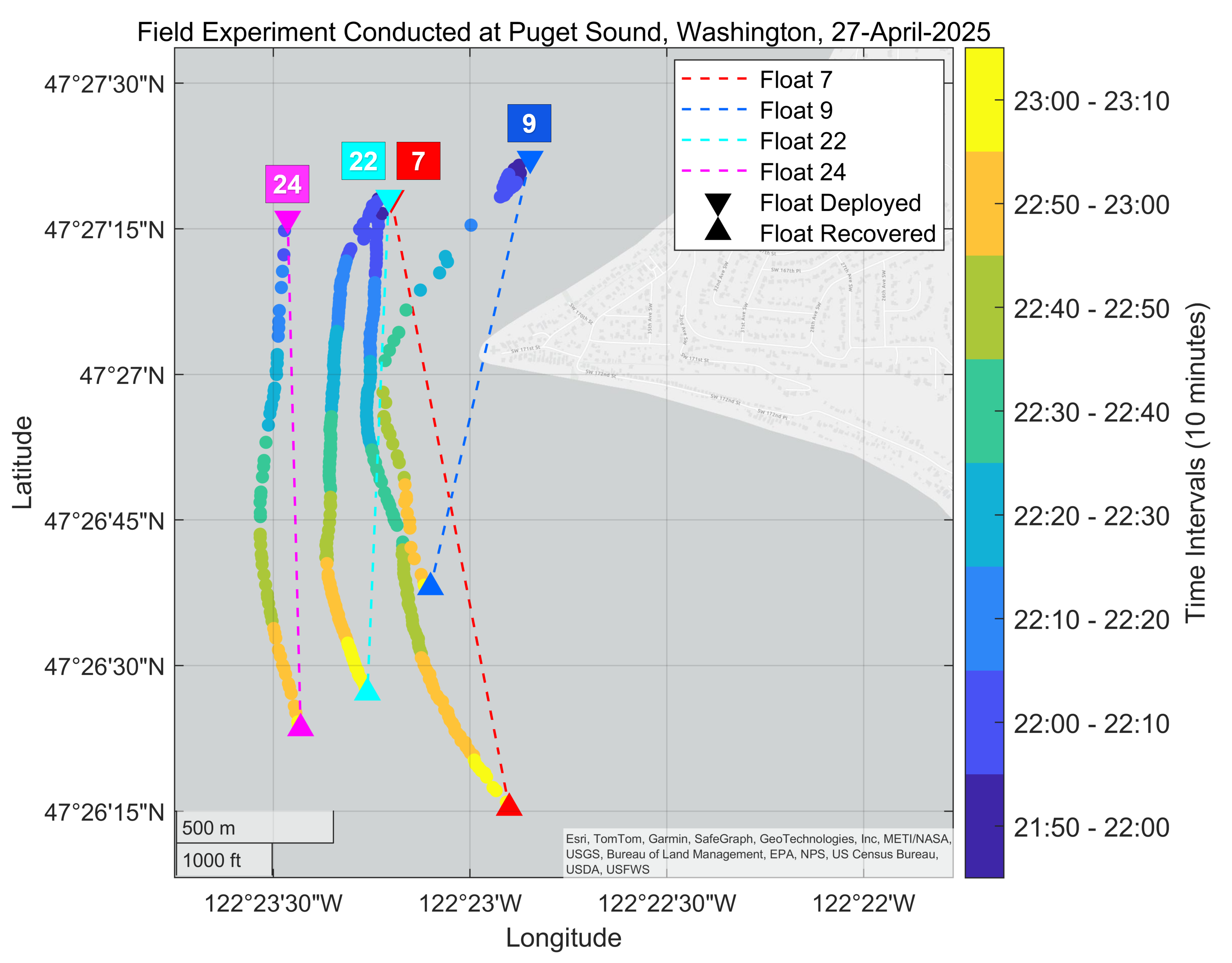}
        \caption{Localization of Float 7, 9, 22, and 24 from Field Deployment II}
        \label{fig:all_float}
    \end{subfigure}
    
    \caption{Localization of the recovered floats from the field deployments.}
    \label{fig:recovered_floats_grid}
\end{figure*}

\begin{figure*}[t!]
    \centering       \includegraphics[width=0.72\linewidth]{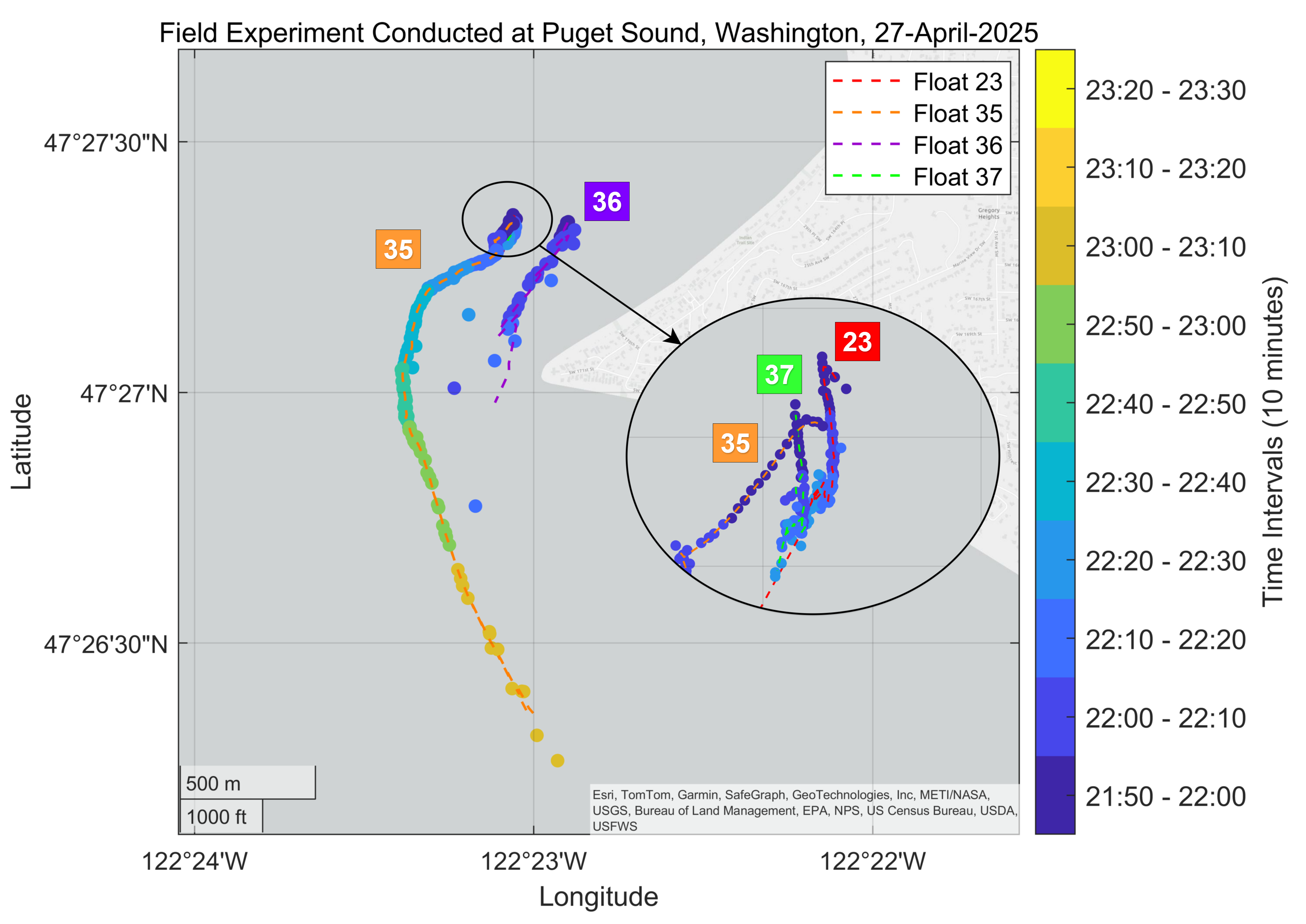}
        \caption{Localization of four of the seven unrecovered floats from Field Deployment II. A moving average filtered trajectory is applied to the position estimates to help distinguish the tracks.}
        \label{fig:unrecovered_floats}
\end{figure*}

\subsection{Localization of Floats}

\subsubsection{Field Deployment I}
In this deployment, a tethered test was conducted on a single float (Float 38), where the float was loosely tethered to SLB 1 to ensure recovery in case of malfunction or leak. The float was scheduled for a 50-minute dive with five 10-minute periods at varying depth levels. All devices were programmed with the same modem schedule, with each device pinging every ten seconds. All data streams were manually reviewed upon recovery for proper operations.

Unfortunately, the float entangled in the tether during this deployment and remained at a shallow depth (less than 20 m) for the duration of the drift. Despite the shallow depth, modem communication with SLBs was consistent for approximately 30 minutes and provided decent data to evaluate the acoustic range estimates. The estimated trajectory of Float 38 is shown in Fig.~\ref{fig:float38}.

\subsubsection{Field Deployment II}

This deployment focused on large-scale float localization. Thirteen floats and the same five SLBs were deployed for approximately one hour. Unfortunately, only six of the thirteen floats (IDs: 7, 9, 22, 24, 39, and 40) were recovered in this deployment, with localizations for four floats shown in Fig.~\ref{fig:all_float}. However, acoustic pings from four of the seven unrecovered floats (23, 35, 36, 37) were received by multiple SLBs and other floats. For these unrecovered devices, we utilized the TDoA algorithm to estimate positions, as shown in Fig.~\ref{fig:unrecovered_floats}.

\section{Discussion}

The results demonstrate that maximizing the use of available acoustic ping data--regardless of transmission direction--can improve the robustness and continuity of float localization. We increased the number of usable range measurements by approximately 50\% of uplink exchanges on average (or single direction, Float-to-SLB exchanges) and reduced the reliance on smoothing functions. The CRLB and cost-based filtering approach proved effective in removing outliers caused by multipath effects and poor geometric configurations, resulting in median errors consistently under 4 m even in dynamic shallow-water environments.

However, several limitations remain. First, sound speed variability remains a challenge, particularly in thermally stratified environments. Although we used a measured average value, accoutning for local sound speed profiles would likely improve accuracy. Second, poor SLB geometry continues to impact localization, and better-distributed SLB arrays reduce positional uncertainty. Additionally, while we assume constant modem delay and negligible clock drift, these assumptions may not always hold for long-duration dives.

\section{Conclusion and future work}

This paper introduces a robust and cost-effective localization framework for the $\mu$Float swarms based on bidirectional ToF ranging. By maximizing measurement opportunities and incorporating CRLB-based uncertainty estimation, our method achieves high-resolution localization using only low-cost acoustic hardware. Field validation confirms the method’s ability to track floats with sub-4-meter accuracy without smoothing. The framework also supports fallback localization using TDoA for unrecovered floats, making it resilient to device loss.

Future efforts will focus on several key directions to improve accuracy, resolution, and adaptability of $\mu$Float localization. Incorporating moored Underwater Localization Beacons (ULBs) with synchronized clocks can improve localization in deeper or thermally stratified environments, especially where surface SLBs are insufficient or acoustically obstructed. Time-synchronized ULBs would also enable more reliable TDoA-based localization and support 3D tracking. The TDoA algorithms can also enable localization in real-time, provided the SLBs can transmit data to the supervising vessel -- they are equipped with 900 MHz RF modems, but the modems are not currently utilized. Lastly, the generalized bidirectional approach, when combined with prior or assumed knowledge about the acoustic environment, can inform structuring of the ping schedule to shift transmissions to those devices with more favorable transmission paths. 


\bibliography{references.bib}

@article{roemmich2009argo,
  title={The Argo Program: Observing the global ocean with profiling floats},
  author={Roemmich, Dean and Johnson, Gregory C and Riser, Stephen and Davis, Russ and Gilson, John and Owens, W Brechner and Garzoli, Silvia L and Schmid, Claudia and Ignaszewski, Mark},
  journal={Oceanography},
  volume={22},
  number={2},
  pages={34--43},
  year={2009},
  publisher={JSTOR}
}

@article{jayne2017argo,
  title={The Argo program: Present and future},
  author={Jayne, Steven R and Roemmich, Dean and Zilberman, Nathalie and Riser, Stephen C and Johnson, Kenneth S and Johnson, Gregory C and Piotrowicz, Stephen R},
  journal={Oceanography},
  volume={30},
  number={2},
  pages={18--28},
  year={2017},
  publisher={JSTOR}
}

@article{arkema2015embedding,
  title={Embedding ecosystem services in coastal planning leads to better outcomes for people and nature},
  author={Arkema, Katie K and Verutes, Gregory M and Wood, Spencer A and Clarke-Samuels, Chantalle and Rosado, Samir and Canto, Maritza and Rosenthal, Amy and Ruckelshaus, Mary and Guannel, Gregory and Toft, Jodie and others},
  journal={Proceedings of the National Academy of Sciences},
  volume={112},
  number={24},
  pages={7390--7395},
  year={2015},
  publisher={National Academy of Sciences}
}

@article{wilkin2017advancing,
  title={Advancing coastal ocean modelling, analysis, and prediction for the US Integrated Ocean Observing System},
  author={Wilkin, John and Rosenfeld, Leslie and Allen, Arthur and Baltes, Rebecca and Baptista, Antonio and He, Ruoying and Hogan, Patrick and Kurapov, Alexander and Mehra, Avichal and Quintrell, Josie and others},
  journal={Journal of Operational Oceanography},
  volume={10},
  number={2},
  pages={115--126},
  year={2017},
  publisher={Taylor \& Francis}
}

@article{siiria2019applying,
  title={Applying area-locked, shallow water Argo floats in Baltic Sea monitoring},
  author={Siiri{\"a}, Simo and Roiha, Petra and Tuomi, Laura and Purokoski, Tero and Haavisto, Noora and Alenius, Pekka},
  journal={Journal of Operational Oceanography},
  volume={12},
  number={1},
  pages={58--72},
  year={2019},
  publisher={Taylor \& Francis}
}

@incollection{liu2015introduction,
  title={Introduction to coastal ocean observing systems},
  author={Liu, Yonggang and Kerkering, Heather and Weisberg, Robert H},
  booktitle={Coastal Ocean Observing Systems},
  pages={1--10},
  year={2015},
  publisher={Elsevier}
}

@article{stutters2008navigation,
  title={Navigation technologies for autonomous underwater vehicles},
  author={Stutters, Luke and Liu, Honghai and Tiltman, Carl and Brown, David J},
  journal={IEEE Transactions on Systems, Man, and Cybernetics, Part C (Applications and Reviews)},
  volume={38},
  number={4},
  pages={581--589},
  year={2008},
  publisher={IEEE}
}

@inproceedings{jiang2007asymmetric,
  title={An asymmetric double sided two-way ranging for crystal offset},
  author={Jiang, Yi and Leung, Victor CM},
  booktitle={2007 International Symposium on Signals, Systems and Electronics},
  pages={525--528},
  year={2007},
  organization={IEEE}
}

@article{zhang2020efficient,
  title={Efficient underwater acoustical localization method based on time difference and bearing measurements},
  author={Zhang, Liang and Zhang, Tao and Shin, Hyo-Sang and Xu, Xiang},
  journal={IEEE Transactions on Instrumentation and Measurement},
  volume={70},
  pages={1--16},
  year={2020},
  publisher={IEEE}
}

@inproceedings{kinsey2006survey,
  title={A survey of underwater vehicle navigation: Recent advances and new challenges},
  author={Kinsey, James C and Eustice, Ryan M and Whitcomb, Louis L},
  booktitle={IFAC conference of manoeuvering and control of marine craft},
  volume={88},
  pages={1--12},
  year={2006},
  organization={Lisbon}
}

@article{Harrison2023benchmark,
author = {Harrison, Trevor and Crisp, Corey and Noe, Jessica and Joslin, James B and Riel, Cassie and Dunbabin, Matthew and Neasham, Jeffrey and Mundon, Timothy R and Polagye, Brian},
title = {Adaptable Distributed Sensing in Coastal Waters: Design and Performance of the $\mu$Float System},
journal = {Field Robotics},
volume = {2},
pages = {516-543},
year = {2023},
DOI = {https://doi.org/10.55417/fr.2023016},
}

@techreport{van1968detection,
  title={Detection and estimation theory},
  author={Van Trees Jr, Harry L and Baggeroer, Arthur B and Collins, LD and Kurth, RR and Cruise, TJ},
  year={1968},
  institution={Research Laboratory of Electronics (RLE) at the Massachusetts Institute of~…}
}

@book{kay1993fundamentals,
  title={Fundamentals of statistical signal processing: estimation theory},
  author={Kay, Steven M},
  year={1993},
  publisher={Prentice-Hall, Inc.}
}

@inproceedings{norrdine2012algebraic,
  title={An algebraic solution to the multilateration problem},
  author={Norrdine, Abdelmoumen},
  booktitle={Proceedings of the 15th international conference on indoor positioning and indoor navigation, Sydney, Australia},
  volume={1315},
  year={2012}
}

@article{li2016contributed,
  title={Contributed Review: Source-localization algorithms and applications using time of arrival and time difference of arrival measurements},
  author={Li, Xinya and Deng, Zhiqun Daniel and Rauchenstein, Lynn T and Carlson, Thomas J},
  journal={Review of Scientific Instruments},
  volume={87},
  number={4},
  year={2016},
  publisher={AIP Publishing}
}

@article{foy2007position,
  title={Position-location solutions by Taylor-series estimation},
  author={Foy, Wade H},
  journal={IEEE transactions on aerospace and electronic systems},
  number={2},
  pages={187--194},
  year={2007},
  publisher={IEEE}
}

@article{fenucci2018development,
  title={Development of smart networks for navigation in dynamic underwater environments. In 2018 IEEE/OES Autonomous Underwater Vehicle Workshop (AUV)},
  author={Fenucci, Davide and Munafo, Andrea and Phillips, Alexander B and Neasham, Jeffrey and Gold, Naomi and Sitbon, Jeremy and Vincent, Iain and Sloane, Terry},
  journal={IEEE, https://doi. org/10.1109/AUV},
  year={2018}
}

@article{pawlak2003observations,
  title={Observations on the evolution of tidal vorticity at a stratified deep water headland},
  author={Pawlak, G and MacCready, P and Edwards, KA and McCabe, R},
  journal={Geophysical Research Letters},
  volume={30},
  number={24},
  year={2003},
  publisher={Wiley Online Library}
}

\end{document}